\documentclass[conference]{IEEEtran}
\IEEEoverridecommandlockouts
\usepackage{cite}
\usepackage{amsmath,amssymb,amsfonts}
\usepackage{algorithmic}
\usepackage{graphicx}
\usepackage{textcomp}
\usepackage{xcolor}
\usepackage{array}
\usepackage{amsmath}
\usepackage{booktabs}
\usepackage{multirow}
\usepackage{multicol}
\usepackage{hyperref}
\definecolor{customPink}{RGB}{238, 193, 255}
\definecolor{lightGray}{RGB}{200, 200, 200}
\newcolumntype{P}[1]{>{\centering\arraybackslash}p{#1}}
\newcolumntype{M}[1]{>{\centering\arraybackslash}m{#1}}

\def\BibTeX{{\rm B\kern-.05em{\sc i\kern-.025em b}\kern-.08em
    T\kern-.1667em\lower.7ex\hbox{E}\kern-.125emX}}
\begin{document}

\title{Live Demonstration: Neuromorphic Radar for Gesture Recognition\\
\vspace{0.5em}
\small IEEE International Conference on Acoustics, Speech and Signal Processing 2025 (ICASSP 2025)\\
Show and Tell Demonstration: 11th April 2025
}

\author{
\IEEEauthorblockN{
Satyapreet Singh Yadav$^{\star}$, 
Akash K S$^{\star}$, 
Chandra Sekhar Seelamantula$^{\dagger}$, 
Chetan Singh Thakur$^{\star}$
}
\IEEEauthorblockA{
$^{\star}$Department of Electronic Systems Engineering, Indian Institute of Science, Bangalore, India 560012\\
$^{\dagger}$Department of Electrical Engineering, Indian Institute of Science, Bangalore, India 560012\\
\{satyapreets, css, csthakur\}@iisc.ac.in, akash2001ks@gmail.com
}
}

\maketitle

\begin{figure}[h!]
    \centering
    \includegraphics[width=0.5\textwidth]{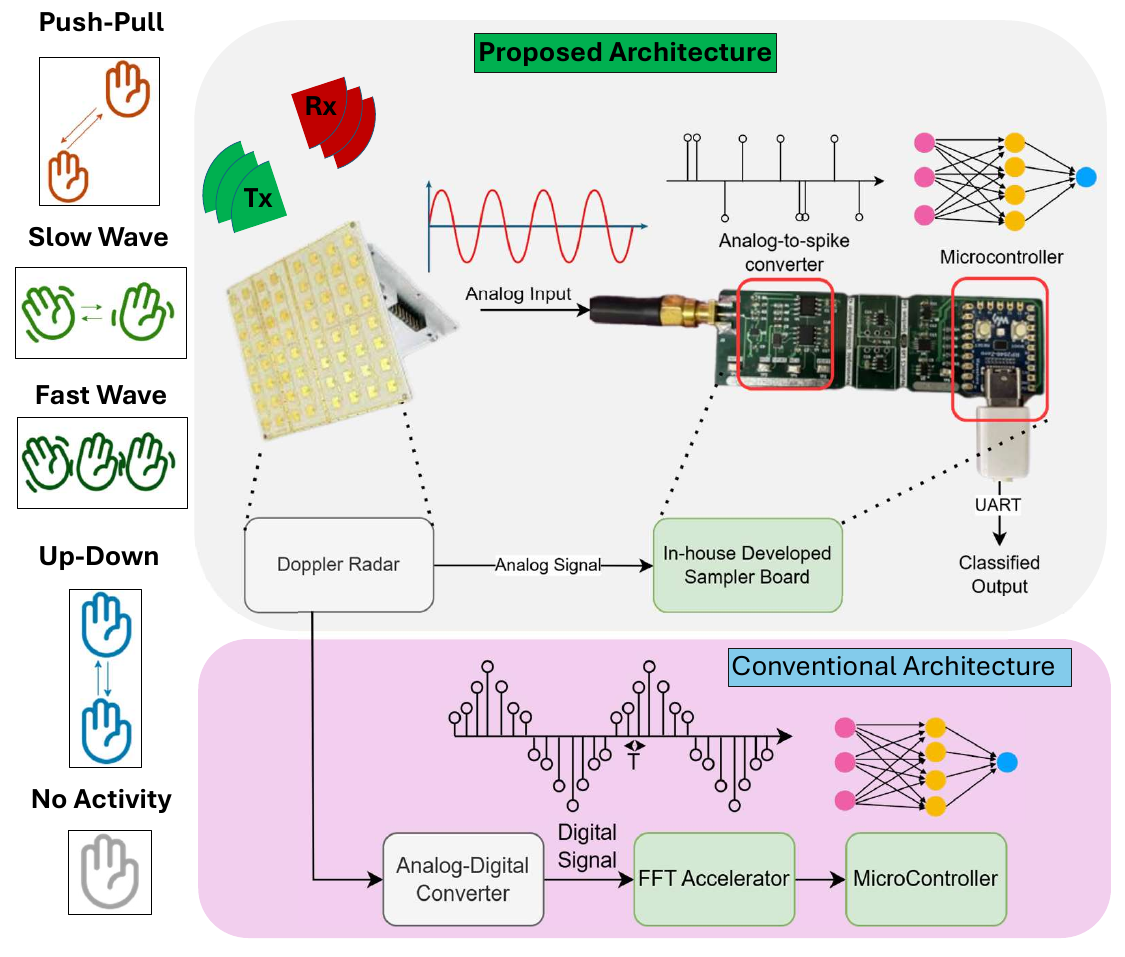}
    \caption{Block diagram of the Neuromorphic Radar Gesture Recognition Demo. The system recognises five distinct gestures: Push-Pull, Slow Wave, Fast Wave, Up-Down, and No Activity. The neuromorphic radar setup includes an analogue pulse-doppler radar, a neuromorphic sampler board that converts IF signals into bio-analogous spike-based representations, and a neural network deployed on a Cortex-M0 microcontroller for real-time gesture inference. The Neuromorphic HGR framework is shown in \textcolor{lightGray}{Grey}, and the conventional architecture using ADC is shown in \textcolor{customPink}{Pink}.
}
    \label{mainDiagram}
\end{figure}

\section{Demonstration Setup}
Our demonstration showcases a neuromorphic radar system designed for real-time hand gesture recognition (HGR), as illustrated in Figure~\ref{mainDiagram}. The setup consists of an analog Doppler radar front-end connected to a custom neuromorphic sampler board that converts intermediate-frequency (IF) signals into sparse, spike-based representations inspired by biological sensing. These neuromorphic signals are then fed into a lightweight neural network deployed on an ARM Cortex-M0 microcontroller \cite{armcortexm0}, enabling event-driven inference.

The primary objective of this setup is to demonstrate real-time classification of five distinct hand gestures: Push-Pull, Slow Wave, Fast Wave, Up-Down, and No Activity. During the demonstration, a user performs these gestures within the radar’s detection zone. The analogue radar captures the motion as Doppler shifts, which are then transformed into spikes by the neuromorphic sampler. This transformation mimics biological sensory encoding, allowing the system to operate efficiently with minimal data.

The neural network processes the incoming spike streams in real time to infer the performed gesture. The output gesture label is transmitted to a connected PC via UART. A Python script on the PC interprets the gesture class and maps it to intuitive controls for a YouTube video interface—such as volume adjustment, play/pause, mute, and seek operations.

This neuromorphic radar setup offers a biologically inspired, data-efficient, and power-conscious alternative to conventional radar systems. By observing the system in action, users can experience how neuromorphic sensing and computing enable real-time, interactive gesture recognition on resource-constrained embedded platforms. To the best of our knowledge, this is the first work that employs bio-inspired asynchronous sigma-delta encoding and an event-driven processing framework for radar-based HGR.

\section{System Description}

Figure~\ref{mainDiagram} illustrates the operation of the neuromorphic radar system. It employs a pulse-doppler radar that transmits a 24\,GHz sinusoidal waveform using a single transmit antenna. When a user performs a hand gesture, the transmitted signal is reflected by the moving hand and received by the receive antenna. The received signal is then down-converted to an IF signal.

This IF signal is fed into our in-house neuromorphic sampler board, which generates spike-based (event-driven) outputs. These events encode the Doppler-induced motion information in a sparse format. The event generation follows the asynchronous sigma-delta encoding scheme described below:

\[
V_{\text{out}} = 
\begin{cases}
+1, & \text{if } V_{\text{in}}(t = \tau) > V_{\text{ref}} \\
-1, & \text{if } V_{\text{in}}(t = \tau) < V_{\text{ref}}
\end{cases}
\]

\noindent
where \( \tau \) is the time when a spike or event is generated. After each event, the reference voltage \( V_{\text{ref}} \) is updated to the current input voltage \( V_{\text{in}}(t = \tau) \), ensuring that future events are generated relative to the most recent signal value. This is a bio-inspired temporal encoding scheme where positive and negative events (spikes) represent upward or downward changes in signal amplitude beyond a certain threshold.  It mimics the behaviour of certain sensory neurons, such as those in the retina or auditory system, which emit spikes only when there is a significant change in stimulus, thereby focusing attention on meaningful variations in the input \cite{lakshmi2019neuromorphic, liu2014event, boahen2025bio}.

The generated events (spikes) are read by the ARM Cortex-M0 microcontroller using an interrupt-driven mechanism, wherein each spike triggers an interrupt and is processed immediately upon occurrence. This scheme is highly efficient for event-based systems, as it eliminates the need for continuous polling and allows the microcontroller to remain in a low-power state between events, saving battery-life. The M0 microcontroller operates at a 125~MHz clock frequency, enabling precise timestamping of each event. Both the spike polarity and its associated timestamp are recorded and used as input features to a lightweight neural network, which performs real-time inference on the embedded platform. This architecture leverages the sparsity and temporal precision of neuromorphic signals to enable fast and efficient gesture recognition with minimal computational overhead.

\section{Novelty of Neuromorphic Radar Framework}
A typical HGR framework using a conventional radar system is illustrated in Figure~\ref{mainDiagram}. It includes an analog-to-digital converter (ADC) that samples the incoming IF signal at a high rate. The digitized data is stored in a buffer and then processed using a hardware Fast Fourier Transform (FFT) accelerator. This processing typically generates feature maps such as doppler-time, range-doppler, or range-doppler-angle representations. These maps provide distinctive motion signatures that are subsequently fed into a classification model, usually a convolutional neural network (CNN) or a recurrent neural network (RNN). Such architectures have been widely adopted in state-of-the-art HGR systems, particularly for Internet-of-Things (IoT) applications \cite{10509948, yadav2023live, wang2016soli}.

However, gesture recognition is inherently a sparse and event-driven task. In most real-world scenarios, the user is not continuously performing gestures. Despite this, the ADC continues to sample the incoming signal, which often contains only background noise. The entire signal acquisition, storage, and processing pipeline remains active, irrespective of whether any meaningful motion is present. As a result, the system processes large volumes of irrelevant data, and the neural network is forced to classify noisy spectrograms merely to conclude that no gesture was performed. This continuous, blind processing imposes a significant burden on memory, power consumption, and latency, making it inefficient for always-on, low-power embedded systems.

The neuromorphic radar framework offers a fundamentally different and efficient approach to gesture recognition through its event-driven architecture. Unlike conventional systems that continuously sample and process signals regardless of activity, the neuromorphic sampler generates positive and negative spikes \textit{only} when a meaningful gesture is performed or when there is a significant change in the IF signal compared to background noise.

This event-driven behaviour is supported by an interrupt-based processing pipeline, wherein the microcontroller samples and timestamps spikes only as they occur. Consequently, there is no need for full signal reconstruction or computation of dense spectrograms. Instead, the system directly utilises the \textit{polarity} and \textit{precise timing} of the spikes to perform real-time inference using a lightweight neural network.

By eliminating unnecessary sampling, processing, and storage during idle periods, the neuromorphic radar framework achieves substantial savings in power, memory, and latency. This makes it especially well-suited for low-power, always-on gesture recognition applications in embedded and IoT systems.

\begin{figure}[h!]
    \centering
    \includegraphics[width=0.4\textwidth]{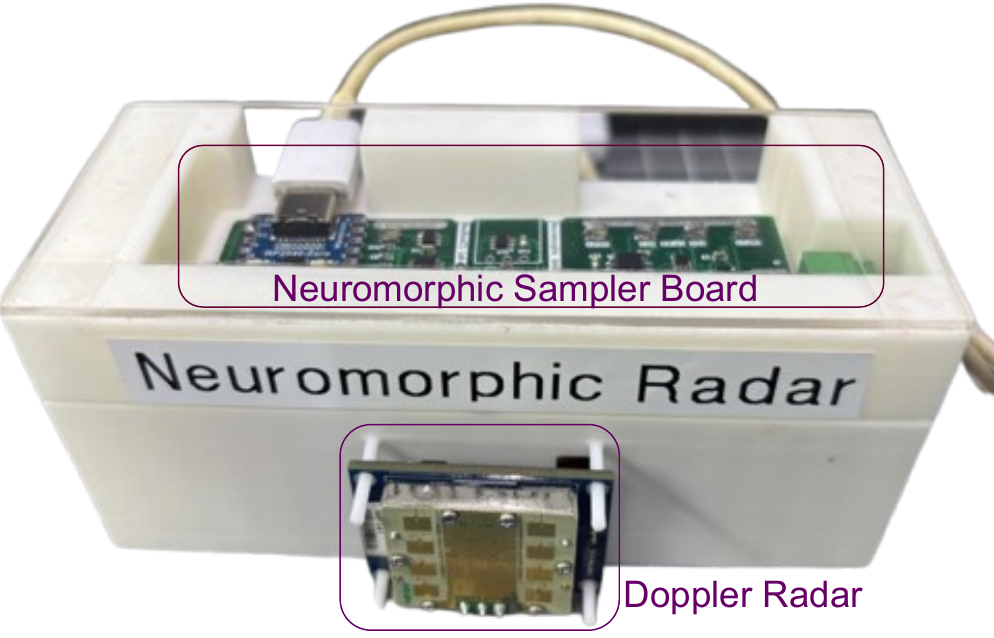}
    \caption{Hardware setup of the Neuromorphic HGR system. It consists of a 24\,GHz pulse Doppler radar and a custom in-house neuromorphic sampler board integrated with a Cortex-M0 microcontroller.}
    \label{Hardware}
\end{figure}

\section{Results}
The neuromorphic radar is a real-time, event-driven gesture recognition solution that utilises a Doppler radar front-end and a custom neuromorphic sampler for bio-inspired encoding of motion, as shown in Figure~\ref{Hardware}.  The gesture recognition model is implemented as a lightweight neural network that operates directly on the polarity and timestamps of the spikes, eliminating the need for spectrogram reconstruction or dense feature maps. The model has a memory footprint of approximately 4\,KB.

We evaluated our system on a custom dataset comprising five gestures, (1) Push-Pull, (2) Slow Wave, (3) Fast Wave, (4) Up-Down, and (5) No Activity, collected from 7 users. Our framework achieves an average real-time inference accuracy of $\geq$\,85\%. The neuromorphic pipeline significantly reduces data movement and processing latency, demonstrating the feasibility of always-on, embedded gesture recognition without compromising accuracy. To the best of our knowledge, this is the first work that employs bio-inspired asynchronous sigma-delta encoding and an event-driven processing framework for radar-based HGR. A live demonstration of our system can be viewed at: \href{https://www.youtube.com/watch?v=1vnsPwD9kqo}{\textit{\textcolor{blue}{Neuromorphic Radar HGR demo video}}}.

\section*{Acknowledgment}
This work originated from discussions held during the Bangalore Neuromorphic Engineering Workshop 2025 (BNEW2025). The authors thank the organising committee for facilitating this event.

 \bibliographystyle{IEEEtran}
 \bibliography{refs.bib}

\end{document}